\documentclass[journal]{IEEEtran}

\ifCLASSOPTIONcompsoc

  \usepackage[nocompress]{cite}
\else
  \usepackage{cite}
\fi

\usepackage{times}
\usepackage{epsfig}
\usepackage{graphicx}
\usepackage{amsmath}
\usepackage{amssymb}
\usepackage{xcolor}
\usepackage{textcomp}
\usepackage{multirow}
\usepackage{mathtools}
\usepackage{breqn}
\usepackage{kantlipsum}
\usepackage{algorithm}
\usepackage{algpseudocode}
\usepackage{tabularx}
\usepackage{subfigure}

\usepackage[algo2e]{algorithm2e}
\usepackage{array}
\newcolumntype{x}[1]{>{\centering\arraybackslash\hspace{0pt}}p{#1}}
\newcolumntype{y}[1]{>{\lecentering\arraybackslash\hspace{0pt}}p{#1}}
\algnewcommand\algorithmicforeach{\textbf{for each}}
\algdef{S}[FOR]{ForEach}[1]{\algorithmicforeach\ #1\ \algorithmicdo}

\newcolumntype{Y}{>{\centering\arraybackslash}X}
\usepackage{caption}

\usepackage{listings}
\usepackage{xcolor}

\definecolor{codegreen}{rgb}{0,0.6,0}
\definecolor{codegray}{rgb}{0.5,0.5,0.5}
\definecolor{codepurple}{rgb}{0.58,0,0.82}
\definecolor{backcolour}{rgb}{0.95,0.95,0.92}

\lstdefinestyle{mystyle}{
    backgroundcolor=\color{backcolour},   
    commentstyle=\color{codegreen},
    keywordstyle=\color{magenta},
    numberstyle=\tiny\color{codegray},
    stringstyle=\color{codepurple},
    basicstyle=\ttfamily\footnotesize,
    breakatwhitespace=false,         
    breaklines=true,                 
    captionpos=b,                    
    keepspaces=true,                 
    frame=single,
    numbersep=5pt,                  
    showspaces=false,                
    showstringspaces=false,
    showtabs=false,                  
    tabsize=2,
}
\usepackage{scalerel,stackengine}

\stackMath
\newcommand\reallywidehat[1]{%
\savestack{\tmpbox}{\stretchto{%
  \scaleto{%
    \scalerel*[\widthof{\ensuremath{#1}}]{\kern-.6pt\bigwedge\kern-.6pt}%
    {\rule[-\textheight/2]{1ex}{\textheight}}%WIDTH-LIMITED BIG WEDGE
  }{\textheight}% 
}{0.5ex}}%
\stackon[1pt]{#1}{\tmpbox}%
}
\usepackage[utf8]{inputenc}

\DeclareMathOperator*{\mymax}{max}

\usepackage[breaklinks=true,bookmarks=false]{hyperref}

\begin{document}

%%%%%%%%% TITLE
\title{Information Maximization Clustering via Multi-View Self-Labelling}

\author{Foivos~Ntelemis,
        Yaochu~Jin, \emph{Fellow}, \emph{IEEE},
        Spencer~A.~Thomas
\thanks{This project is funded by an EPSRC industrial CASE award (number 17000013) and Department for Business, Energy and Industrial Strategy through the National Measurement System (122416). (\textit{Corresponding author: Yaochu Jin})}
\thanks{F. Ntelemis and Y. Jin are with the Department of Computer Science, University of Surrey, Guildford, GU2 7XH, United Kingdom. (Email: \{f.ntelemis; yaochu.jin\}@surrey.ac.uk)}
\thanks{S. A. Thomas is with the National Physical Laboratory, Teddington, TW11 0LW, United Kingdom. (Email: spencer.thomas@npl.co.uk)}
%and Technology, Shanghai, 200237, P. R. China (e-mail: \{donghan;duwei0203;wldu\}@ecust.edu.cn).}% <-this % stops a space
%\thanks{J. Doe and J. Doe are with Anonymous University.}% <-this % stops a space
%\thanks{Manuscript received xx, 2021; revised xx, 2021.}
}

\maketitle
% Remove page # from the first page of camera-ready.

\begin{abstract}

Image clustering is a particularly challenging computer vision task, which aims to generate annotations without human supervision. Recent advances focus on the use of self-supervised learning strategies in image clustering, by first learning valuable semantics and then clustering the image representations. These multiple-phase algorithms, however, involve several hyper-parameters and transformation functions, and are computationally intensive. By extending the self-supervised approach, this work proposes a novel single-phase clustering method that simultaneously learns meaningful representations and assigns the corresponding annotations. This is achieved by integrating a discrete representation into the self-supervised paradigm through a classifier net. Specifically, the proposed clustering objective employs mutual information, and maximizes the dependency between the integrated discrete representation and a discrete probability distribution. The discrete probability distribution is derived through the self-supervised process by comparing the learnt latent representation with a set of trainable prototypes. To enhance the learning performance of the classifier, we jointly apply the mutual information across multi-crop views. Our empirical results show that the proposed framework outperforms state-of-the-art techniques with an average clustering accuracy of 89.1\%, 49.0\%, 83.1\% and 27.9\%, respectively, on the baseline datasets of CIFAR-10, CIFAR-100/20, STL10 and Tiny-ImageNet/200. Finally, the proposed method also demonstrates attractive robustness to parameter settings, and to a large number of classes, making it ready to be applicable to other datasets. The implementation of our method is available  \href{https://github.com/foiv0s/imc-swav-pub}{online}.

\end{abstract}


\IEEEdisplaynontitleabstractindextext

\IEEEpeerreviewmaketitle

\section{Introduction}\label{sec:introduction}

Modern technologies such as Internet of Things and cloud computing have resulted in the collection and storage of a huge amount of data such as images and videos. With the help of such huge amount of data, deep supervised learning \cite{7780459,7298594} has achieved great success, provided that these data are labelled. Unfortunately, labelling such huge datasets is extremely laborious, and in many cases intractable. As a result, many image datasets are not fully utilized due to the lack of labels. In addition to the huge volumes, image data are usually characterized by a high dimensionality and multi-modal structure. Thus, the performance of traditional clustering approaches \cite{BEZDEK1984191,DBLP:conf/iccv/ComaniciuM99,Heller2005BayesianHC,Williams1999AMA,10.1007/978-3-642-33718-5_31} seriously deteriorates on such data  \cite{DBLP:conf/iccv/ChangWMXP17,Steinbach2004}. By contrast, deep unsupervised methods have demonstrated superiority and scalability in handling vision data \cite{NIPS2012_c399862d}, including representation learning \cite{caron2020unsupervised,pmlr-v119-chen20j,He_2020_CVPR} and clustering \cite{8237888,DBLP:conf/eccv/HanPPKC20,Ji_2019_ICCV,Ren2020DeepDI,10.1007/978-3-030-58607-2_16}.

Representation learning techniques are gaining popularity, since they can generate discriminative features by considering spatial properties, object shapes and photometric information without the requiring of human annotations. Recent studies \cite{caron2020unsupervised,pmlr-v119-chen20j,He_2020_CVPR,2018arXiv180703748V} have dramatically enhanced the learning capacity and thus minimized the performance gap between supervised and unsupervised learning tasks. This is achieved by proposing self-training objectives for representation learning, also called self-supervised methods.
In particular, contrastive learning strategies have widely been applied in self-supervised methods \cite{ caron2020unsupervised,pmlr-v119-chen20j,pmlr-v9-gutmann10a,2018arXiv180703748V}. They aim to increase the concordance of positive (similar) features of augmented views and decrease the discordance in negative (dissimilar) instances. Alternatively, feature comparisons are replaced in grouping techniques \cite{asano2020self, caron2018deep,caron2020unsupervised} by identifying sub-classes and assigning pseudo-labels to the relevant representations based on their similarities. These pseudo-labelling annotations are either derived through traditional clustering methods \cite{caron2018deep} (i.e., extracted features using a convolutional encoder are annotated through K-means clustering) or through an optimal transport plan \cite{asano2020self,caron2020unsupervised} that can ensure consistency and balance population. 
\par

Despite the remarkable success of grouping based self-supervised tasks, the pseudo-labels they generate have so far been evaluated for learning discriminative features only. Additionally, the deep clustering methods have not yet fully exploited the recently developed self-supervised algorithms. Some existing approaches \cite{Hu2017,Ji_2019_ICCV} implement a convolutional framework that introduces mutual information (MI) in the clustering objective during training. The application of MI has demonstrated to be beneficial; however, the performance of these methods degrades in tackling more challenging datasets. Most recently, self-supervised mechanisms were introduced to further improve clustering results on challenging datasets \cite{DBLP:conf/eccv/HanPPKC20,vangansbeke2020scan}, by firstly learning valuable properties of the training instances and then performing the clustering. Although the performance is enhanced, they often require multiple training phases whose clustering performance relies on the previous training phase and the effectiveness of each individual stage. Furthermore, many individual training phases require an additional set of augmentation operations and/or different objective functions. As a result, the final clustering accuracy heavily relies on an increasing number of hyper-parameters. Finally, multiple training phases often increase the computational complexity. 

This study aims to address the issues of multiple-phase clustering methods by extending the functionality of Swapping Assignments between multiple Views (SwAV) \cite{caron2020unsupervised}, a recently proposed grouping based self-supervised learning strategy that performs discriminative feature learning. Therefore, we propose an online single-phase clustering framework by integrating a deep classifier net into the SwAV framework to simultaneously learn the representations and assign the desired annotations. Specifically, the proposed clustering method exploits the semantic structure obtained through a discrete probabilistic distribution that is derived by comparing the latent representations with a set of trainable prototypes. This is accomplished by maximizing the mutual dependency between the discrete representation of the integrated classifier and the discrete probabilistic distribution derived by the prototypes. To further enhance the clustering performance, we adopt the multi-crop strategy as presented in \cite{caron2020unsupervised} and optimize the mutual information across multi-crop views. We call the proposed framework Information Maximization Clustering by Swapping Assignments between multiple Views (IMC-SwAV). Our empirical studies show that the proposed framework is highly competitive and outperforms a wide range of state-of-the-art methods. Additionally, it shows robustness to the number of prototypes and the number of clusters. Finally, we demonstrate that other grouping-based self-supervised learning methods such as SeLa \cite{asano2020self} can also be adopted, although the proposed single-phase clustering method is based on an extended SwAV. 

We highlight our contributions as follows: 
%1) A single training phase framework for image clustering is proposed by extending the operation of a self-supervised technique to simultaneously assign cluster labels and extract disentangled features; 
%2) A modified clustering objective is proposed based on mutual information, which maximizes the dependency between the discrete representation predicted by the classifier and the probabilistic distribution obtained by the trainable prototypes;
%3) The multi-crop views strategy is adopted in optimizing the classifier to further enhance the classifier's performance.
\begin{enumerate}
    \item A single-phase training framework is proposed by extending a feature learning strategy to simultaneously generated features and assign labels. This way, the proposed algorithm avoids extra hyper-parameters, and does not require additional transformation functions, nor extra training phases.
    \item A modified clustering objective based on mutual information that maximizes the dependency between an over-clustering discrete probabilistic distribution and the classifier's discrete representation. A multi-crop view strategy is adopted in optimizing the classifier to further enhance the classifier's performance.
    \item We demonstrate that our clustering strategy is not limited to a specific self-supervised learning strategy and can in principle be extended to any other grouping based self-supervised learning strategies within a single-phase training framework.
\end{enumerate}
The rest of the paper is organized as follows. In Section II, related work on clustering and representation learning are briefly reviewed. The proposed method, including the overall representation learning framework and a modified objective function for joint training on the basis of mutual information maximization is presented in Section III. Comparative experiments and ablation studies, along with the parameter settings and performance metrics, are described in Section IV. A summary of the proposed algorithm and future work are provided in Section V.    

\section{Related work}

\textbf{Embedding strategies} have emerged as an effective means to eliminate the necessity of the desired annotations in terms of disentangled feature learning. Previous approaches employ traditional frameworks, such as  autoencoders \cite{HintonSalakhutdinov2006b,Vincent2008} and deep belief networks \cite{10.1162/neco.2006.18.7.1527}. Generative models including variational autoencoders \cite{DBLP:conf/aistats/KhemakhemKMH20} and generative adversarial networks \cite{10.5555/2969033.2969125,DBLP:journals/corr/RadfordMC15} also became popular. This type of frameworks rely on two separate models,  one generating training instances from a latent space, and the other mapping the training samples into a latent domain. \par

\textbf{Self-supervised learning}, on the other hand, obtains visual representations using a single unit model. A range of training objectives have been proposed in literature: 1)  each training sample is treated as a unique class \cite{DBLP:journals/pami/DosovitskiyFSRB16}; 2) an image instance is partitioned in several patches and the encoder either attempts to solve the jigsaw puzzle \cite{10.1007/978-3-319-46466-4_5}, or 3) predicts the code of next image's patch \cite{2018arXiv180703748V}. Contrastive learning addresses feature extraction by comparing positive pairs (augmented views of similar instances). Many implementations \cite{caron2020unsupervised,pmlr-v119-chen20j,He_2020_CVPR,2018arXiv180703748V, DBLP:journals/ijcv/ZhaoXWWTL20} adopt variations of noise contrastive estimation (NCE) \cite{pmlr-v9-gutmann10a} loss function, that achieves a good representation by comparing positive pairs versus a large number of negative instances. To deal with the large negative sampling, momentum contrast (MoCo) \cite{He_2020_CVPR} has been suggested that compares representations between a momentum and a simple encoder, or in \cite{DBLP:journals/ijcv/ZhaoXWWTL20} through a memory bank for storing past representations. As another option, the grouping based strategies \cite{asano2020self,caron2018deep,caron2020unsupervised} assign the representations to a numerous surrogate sub-classes. A method presented in \cite{caron2018deep} employs a traditional K-means to cluster representations, whose performance, however, depends on network initialization, and an additional computational time is required for computing the centroids in each iteration. Some recently proposed algorithms \cite{asano2020self,caron2020unsupervised} take advantage of the efficient optimal transport plan, and introduce a self-labelling mechanism as target distribution which is computed via the Shinkhorn-Knopp \cite{NIPS2013_af21d0c9} algorithm. Nevertheless, these methods aim to extract features without producing cluster annotations.

\begin{figure*}[t]
    \centering
    \includegraphics[width=1.\textwidth]{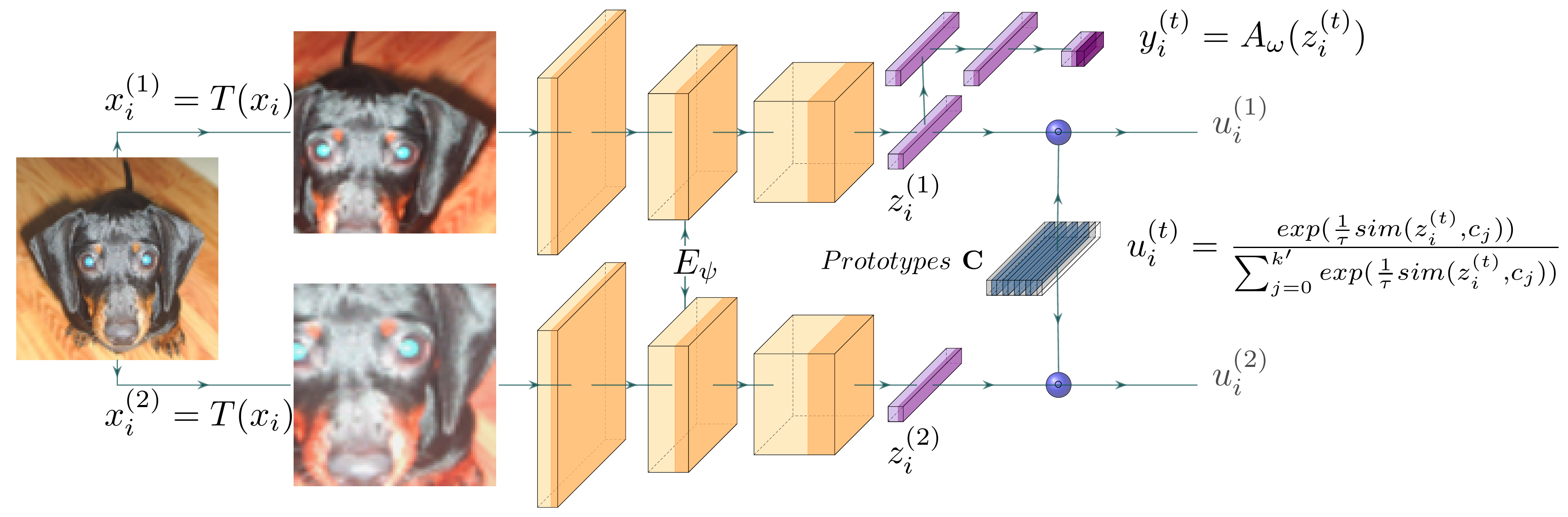}
    \caption{A diagram presents the framework's structure and a training instance $x_i$, transformed twice through $T()$. Here $E_\psi$ denotes the encoder model, and the comparable prototypes as $\textbf{C}$. $A_\omega$ indicates the introduced classification model implemented on top of the embedding output (diagram is designed via PlotNeuralNet \cite{haris_iqbal_2018_2526396}).}
    \label{fig:framework}
\end{figure*}

\par
\textbf{Image clustering} techniques group the corresponding representation basis to the defined priorities. Two main strategies are considered from different perspectives. A first group of studies \cite{DBLP:conf/iccv/ChangWMXP17,Ji_2019_ICCV, Yang2016JointUL,zhao2020deep} apply a defined clustering loss function to identify patterns within the training samples. A method called deep adaptive clustering (DAC) \cite{8237888} implements a convolutional net, and measures the cosine similarities of the generated features. The model is gradually trained to assign the most similar features to the same group index. Ji \textit{et al.} propose invariant information clustering (IIC) \cite{Ji_2019_ICCV} which includes the MI as a training objective to maximize the dependency between the categorical outputs of numerous augmented views. A similar algorithm is reported in \cite{Hu2017}, which suggests a variation of MI as the training objective for learning discrete representations regularized through virtual adversarial training. Nevertheless, the performance of the aforementioned single-phase training methods deteriorate on more challenging datasets. A second group of most recent methods is based on multiple sequential training phases. State-of-the-art performance has reported in \cite{vangansbeke2020scan}, where the model is initialized in the first phase by applying a contrastive learning objective. In the second phase, the $k$ nearest features of each instance are measured and considered to belong to the same group, then the model is trained accordingly. Due to the mismatch prediction of the $k$ nearest features, the method implements a third round of training, where the model learns from the most confident predictions. Likewise, a two-stage clustering is incorporated in \cite{eccv_han}, where the encoder parameters are firstly initialized through a self-supervised strategy. This pre-training phase significantly improves the performance by a large margin of the later clustering objective in the second stage. However, having multiple training stages will no doubt increase the computation time and introduce dependency on previous phases. In addition, each training phase usually involves its specific  hyper-parameters, transformation functions, optimizers as well as objective functions. By contrast, our proposed strategy is based on a single-phase training method, nevertheless, it demonstrates competitive performance comparable to the most recent multiple-phases strategies.

\section{The Proposed Method}
\label{sec:method}
Assume there is an unlabelled set of images denoted as $X = \{x_i\}_{i=1}^{n}$, which holds a relation with a finite set of classes $Y= \{y_i \in \mathbb{N}, 0 < y_i \leq k\}$, where $\textit{k}$ is a given hyper-parameter equal to the number of classes. The goal of this work is to instantiate a convolutional model and map the described relation as $f_{\theta}: X \rightarrow Y$, and $\theta$ indicates the model's parameters. Figure \ref{fig:framework} presents the overall framework of the proposed IMC-SwAV, consisting of three main components: 1) a ConvNet encoder ($E_\psi$) that projects the training instances onto a latent space ($Z$); 2) the trainable prototype vectors $\textbf{C}$ that are compared with the projected features to derive the computed distribution; and 3) an introduced classifier ($A_\omega$), which maps the generated features to the corresponding discrete representation.

In the following, we begin with a brief description of a representation learning strategy based on an online self-labelling assignment method and its application to the optimization of an encoder model. We then present a joint training objective, which is a modified form of mutual information, with the aim to maximise the mutual dependency of the computed distribution obtained by the prototypes and the classifier's predictions.

\subsection{Unsupervised Representation Learning}

The encoder model adopted in this work $E_\psi: X \rightarrow Z$, where $\psi$ denotes the encoder's trainable parameters, aims to learn the important semantic information of the given set of data without supervision, while ignoring less valuable semantics such as background or noise. Hence, motivated by the recent state-of-the-art achievement of SwAV \cite{caron2020unsupervised}, we employ the SwAV contrastive learning strategy to impose the consistency between representations ($Z$) obtained from augmented views of the same instances by comparing them to a set of prototype vectors $\textbf{C}$. These prototypes are evaluated only for uniformly mapping the obtained representations by exchanging their predicted probabilistic distributions, respectively. This is achieved by minimizing the \textit{swapped} self-labelling training objective \cite{caron2020unsupervised}: 
\begin{equation}
\begin{split}
L_{\psi, C}^{\textit{swap}}(z^{(1)},z^{(2)}) &= \ell(z^{(1)}, q^{(2)}) +  \ell(z^{(2)}, q^{(1)}) \\
\text{where  \ \ \ } \ell(z^{(j)}, q^{(l)}) &= -\frac{1}{m} \sum_{i=1}^m q_i^{(j)}log(u_{i}^{(l)})
\label{eq:swav objective}
\end{split}
\end{equation}
where $z_{i}^{(t)}=E_{\psi}(T(x_{i}))$ denotes a latent representation derived by the encoder ($E_\psi$) of the $i\text{-th}$ training instance, respectively, $T \in \{f_1,f_2,..,f_n\}$ is a collection of stochastic transformation functions applied on the original instance beforehand to obtain a transformed view $x_i^{(t)}=T(x_i)$, where $t$ is the augmented view index. For example, representations obtained by two alternative views of the same instance ($x_i$) are indicated as $z_i^{(1)}$ and $z_i^{(2)}$ in Equation \ref{eq:swav objective}, as illustrated in Fig. \ref{fig:framework}. $q_i^{(1)}$ and $q_i^{(2)}$ are the swapped self-labelling target distributions of the two transformed views of the $i$-th training instance, and $u_i^{(1)}$ and $u_i^{(2)}$ are the computed probabilistic distributions, respectively, also called ``codes", determined by comparing the corresponding representations with the trainable prototype vectors $\textbf{C}$ as follows:
\begin{equation}
u^{(t)}_i = \frac{\text{exp}(\frac{1}{\tau}\text{sim}(z_{i}^{(t)},c_j))}{\sum_{j=1}^{k'}\text{exp}(\frac{1}{\tau}\text{sim}(z_{i}^{(t)},c_j))}
\label{eq:u definition}
\end{equation}
where $\tau$ is a temperature parameter that controls the smoothness of the probabilistic Softmax output, and $k'$ indicates the number of prototypes. Lastly, we define $\text{sim}(z_{i},c_j)=z_{i}^T c_j/\|z_{i}\|\|c_j\|$ as the cosine similarity distance of both $\ell_2$ normalized vectors, and $c_j$ represents the $j\text{-th}$ prototype vector. \par

To minimize the cross-entropy terms of the \textit{swapped} training objective (Equation \ref{eq:swav objective}), the computation of the exchanged pair distributions ($q^{(t)}_i$) for each augmented view is required. The self-labelling target distributions ($q^{(t)}_i$) prevent the assignment of trivial solutions, and enforce a uniform mapping to the prototypes. Ultimately, we accommodate the online computation as introduced in \cite{caron2020unsupervised} to derive the required distribution. Given a collection of feature vectors $Z=[z_1,z_2,...,z_m]$, with $m$ being the size of the training mini-batch, and the prototype vectors $C=[c_1,c_2,...,c_{k'}]$, we aim to define a target distribution $Q=[q_1,q_2,...,q_m]$ to maximize the correlation between the generated feature vectors and the trainable prototypes by satisfying the aforementioned conditions as:
\begin{equation}
\mymax\limits_{\bold{Q} \in \mathcal{Q}}  \ \text{Tr}(\bold{Q^{T}C^{T}Z}) + \epsilon H(\bold{Q})
\label{eq:max objective}
\end{equation}
\begin{equation}
\mathcal{Q}  = \{\bold{Q} \in \mathbb{R}_{+}^{k'\times m}| \bold{Q}1_M = \frac{1}{K'}1_{K'}, \bold{Q}^T1_{K'}=\frac{1}{M}1_M\}
\label{eq:q definition}
\end{equation}
where $\epsilon$ is a scaling parameter that regularizes the entropy term $H(\bold{Q})$ by controlling the mapping of the distribution. In practice, the required target distribution can be efficiently optimized by iteratively employing the Sinkhorn-Knopp \cite{NIPS2013_af21d0c9} algorithm. Hence, the constraints of a uniform mapping and homogeneity are encouraged within the training mini-batch. The reader is referred to \cite{caron2020unsupervised} and \cite{NIPS2013_af21d0c9,peyre2020computational}, respectively, for a more detailed description of the SwAV  and Sinkhorn-Knopp algorithms.

\subsection{Joint Cluster Representations}

Thus far, the described encoder generates features without producing the desired annotation or discrete properties to the given $k$ classes. Instead, the training set is mapped in domain $U=\{U \in \mathbb{R}^{m \times k'}_+| \sum_{j=1}^{k'} u_{i,j}=1  \}$. Our goal is to define a parametric classifier that assigns the generated representations into a given number of classes $\textit{k}$ such as $A_\omega : Z \rightarrow Y $, where $\omega$ are parameters of classifier $A$.\par

\begin{figure}[t]
    \centering
    \includegraphics[width=0.4\textwidth]{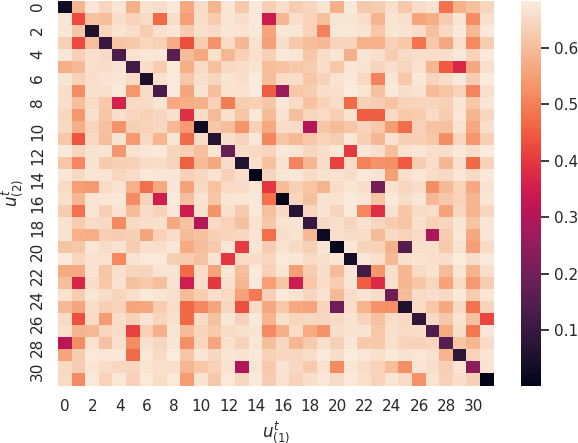}
    \caption{A scatter illustration of JSD pairwise distances between two probabilistic distributions $U^{(1)}$ and $U^{(2)}$ generated by a pre-trained encoder of the same 32 image instances, where different transformations applied in each instance.
    }
    \label{fig:scatter similarities}
\end{figure}
To this end, the encoder is optimized by assigning pairs of the same instances to their corresponding prototypes, while ensuring different instances, that are distinct for each prototype. We evaluate this argumentation by computing two distributions of a mini-batch denoted as $U^{(1)}$ and $U^{(2)}$ using a pre-trained encoder. Each distribution includes the same 32 instances from the training set, where each instance is alternated with the application of the transformation functions $T()$. We measure the pairwise similarity of the two probabilistic distributions by applying the Jensen–Shannon divergence (JSD) \cite{1365067}, a symmetric bounded distance measurement for evaluating two probability distributions. As presented in Fig. \ref{fig:scatter similarities}, the diagonal positions express the lower divergences with the mean value $\approx 0.06$ as these transformed views are from the same original instances. On the other hand, the larger divergences are observed in the instances of different indices (off diagonals) with the mean value $\approx 0.63$. Note that the upper bound of JSD is $log(2) \approx 0.69$ and zero the lower bound. \par

Complementary to the above observation, we assume that the transformed views of an instance produce similar distribution in domain $U$. We can also assume that the probabilistic output ($U$) in Eq. \ref{eq:u definition} and the corresponding classifier's outputs $Y = A_\omega(Z)$ holds a relationship since both are obtained from representations $Z$, hence defining a training objective that \textit{maximizes the mutual dependency} between these two distributions with respect to parameters $\omega$. Then classifier $A_\omega$ will map semantically similar instances in domain $U$ to the same cluster in domain $Y$. Note that the distributions of $Y$ and $U$ are a product of the Softmax output; hence, both belong to the %same
probability simplex. Inspired by \cite{Ji_2019_ICCV}, we propose a measurement describing the amount of dependency between the two probabilistic outputs, which is the direct application of mutual information (MI) \cite{Cover:1991:EIT:129837}:
\begin{equation}
\begin{split}
I(U;Y)  &= - H(Y|U) + H(Y) \\
&= \sum_{(u,y)} p(u,y) log \left (\frac{p(u,y)}{p(u)} \right ) -\sum_{y} p(y)  log (p(y))
\label{eq:mi}
\end{split}
\end{equation}
The MI expression is the relative entropy for measuring the divergence of the joint probability $p(u,y)$ and the product of two marginals. We modify Equation \ref{eq:mi} to define our clustering objective function with respect to parameters $\omega$ of the classifier. In this work, we employ a deep net as the classifier $A_{\omega}$ on top of the embedding layer ($Z$) to jointly perform the training. We invert the corresponding expression into a minimization loss function, and optimizing the model using the stochastic gradient descent:
\begin{equation}
\begin{split}
L^{\textit{cluster}}_\omega (Y,U)  &= \frac{1}{V^2}\sum_{i}^V \sum_{j}^V \left [ H(P_{Y^{j}}|P_{U^i}) - \beta H(P_{Y^{j}}) \right] \\
&= \frac{1}{V^2}\sum_{i}^V \sum_{j}^V \left[ \sum_{Y^j, U^i} P_{Y^j U^i } log \left( \frac{P_{U^i}}{P_{Y^{j} U^{i}}} \right) \right . \\ &+\left . \beta \sum_{Y} P_{Y^j} log(P_{Y^j})\right] ~.
\label{eq:mi v2}
\end{split}
\end{equation}

\begin{lstinputlisting}[language=Python,label={lst: psedocode},float=bp, caption=Pseudo-code written in Python presents the clustering objective based on Pytorch library \cite{NEURIPS2019_9015}.]{sup_includes/psedocode.py}
\end{lstinputlisting}

where $P_{Y^j U^i} = \{\frac{1}{m} {Y^j}^{\!\!\top} U^i \in \mathbb{R}_+^{k \times k'}| Y^j \in \mathbb{R}_+^{m \times k}, U^i \in \mathbb{R}_+^{m \times k'} \}$, $U^i$ denotes the computed codes, $Y_j$ the classifier's prediction, $P_{U^i}= \{\frac{1}{m} \sum^m_{l=1} u^i_{{l,k'}} \in \mathbb{R}^{k'}_{+}\}$ and $P_{Y^j}= \{\frac{1}{m} \sum^m_{l=1} y^j_{{l,k}} \in \mathbb{R}^{k}_{+}\}$ the two marginal terms. Here, $m$ indicates the number of elements in the mini-batch and indices $i$ and $j$ denote the corresponding transformed views of the mini-batch, as illustrated in Fig. \ref{fig:framework}, where $V$ indicates the total number of transformed views obtained from instances in the mini-batch. The scalar $\beta$ denotes an introduced weight parameter of the entropy term $H(P_{Y^j})$, that  encourages the classifier to uniformly assign the predicted class indices and thus prevent degeneracy. The conditional entropy term is minimized by increasing the classifier's prediction confidence. In Listing \ref{lst: psedocode}, we provide the pseudo-code of the clustering objective function as described in Eq. \ref{eq:mi v2}, which returns a quantitative measurement of the dependency between the classifier's predictions ($Y^j$) and the distribution ($U^i$). The pseudo-code is written in Python and includes commands extended in the Pytorch library \cite{NEURIPS2019_9015}. The final training objective with respect to all parameters of the overall framework ${\{\psi,\omega,\textbf{C}\}} \in \theta$ is defined as follows: 
\begin{equation}
\begin{split}
\textit{L}_\theta = \textit{L}_{\psi,C}^{\textit{swap}} + L^{\textit{cluster}}_\omega  ~.
\label{eq:full objective}
\end{split}
\end{equation}
The internal computation of the IMC-SwAV is presented in Algorithm \ref{alg:framework} and the full training process, including all steps for the two transformed views, are given in Algorithm 2.

\SetKwInOut{Initialize}{Initialize}
\SetKwInOut{Generate}{Generate}

\begin{algorithm}[t]
\SetKwFunction{FMain}{Framework}
    \SetKwProg{Fn}{Function}{:}{}
    \Fn{\FMain{$B$}}{
        \hspace{0.1mm} $B^{t} = T(B)$ apply transformation functions\\
        \hspace{0.1mm} $Z^{t}=E_\psi(B^{t})$ generate the representations\\
        \hspace{0.1mm} $U^{t}= \frac{\text{exp}(\frac{1}{\tau}\text{sim}(z_{i}^{t},c_j))}{\sum_{j=1}^{k'}\text{exp}(\frac{1}{\tau}\text{sim}(z_{i}^{t},c_j))}$ compute the distribution \\
        \hspace{0.1mm} $Y^{t} = A_\omega(Z^{t})$ classifier's output \\
        \hspace{0.1mm} $Q^{t} = $  derived by $U^{t}$ through Shinkhorn-Knopp \cite{NIPS2013_af21d0c9}\\
 \Return $U^{t}$ ,  $Q^{t}$ ,  $Y^{t}$
}
\textbf{End Function}

\caption{Framework method}
\label{alg:framework}
\end{algorithm}

\begin{algorithm}[t]
\KwData{$X=\{x_{i}\}_{i=1}^{n}$}
\Initialize{$\{\psi, \omega,\textbf{C} \} \in \theta$}
\While{Convergence condition not satisfied}{
  \hspace{0.7mm}{ \textbullet \hspace{0.1mm} Sample minibatch $B=\{x_{1},...,x_{m}\}$} \\
  \hspace{1.mm}{ \Comment{First Views} }\\
  \hspace{0.7mm}{ \textbullet \hspace{0.1mm} $U^{(1)}$ ,  $Q^{(1)}$ ,  $Y^{(1)} =$  \FMain{$B$} }\\ 
  \hspace{1.mm}{ \Comment{Second Views} }\\
  \hspace{0.7mm}{ \textbullet \hspace{0.1mm} $U^{(2)}$ ,  $Q^{(2)}$ ,  $Y^{(2)} =$  \FMain{$B$}} \\ \\
   \hspace{1.mm}{ \Comment{Swap loss} }
$$
\textit{L}_{\psi, C}^{\textit{swap}} = \frac{1}{m} \sum_{i=1}^m \left ( -q_i^{(2)}log(u_{i}^{(1)}) - q_i^{(1)}log(u_{i}^{(2)}) \right )
$$ \\ 
\hspace{1.mm}{ \Comment{Clustering loss} }\\
%\hspace{0.7mm}{ $U \in \{U^{(1)},U^{(2)}\}$} \\
%\hspace{0.7mm}{ $Y \in \{Y^{(1)},Y^{(2)}\}$} %\\
$$L^{\textit{cluster}}_\omega (U,Y)  = \frac{1}{2^2} \sum_{i=1}^{2} \sum_{j=1}^{2} \left [ H(Y^j|U^i) - \beta H(U^i) \right] $$

\textbullet \hspace{0.1mm} Jointly update the framework's  parameters to minimize the combined training objective $\textit{L}_\theta$: \\
\begin{align*}
%\begin{split}
\textit{L}_\theta = \textit{L}_{\psi,C}^{\textit{swap}} + L^{\textit{cluster}}_\omega
%\end{split}
\end{align*}
}
\caption{The proposed joint training algorithm.}
\label{alg:Process}
\end{algorithm}

In the following, we briefly discuss the reason for using mutual information in the objective function. Recall that the convergence of the SwAV \cite{caron2020unsupervised} framework is achieved by gradually dividing the full dataset into $k'$ partitions, where $k'$ is the number of defined prototypes, which is usually large. Thus, on average $n/k'$ instances are assigned to each partition. By maximizing the dependency between the distribution $U=\{U \in \mathbb{R}^{m \times k'}_+| \sum_{j=1}^{k'} u_{i,j}=1  \}$ and the classifier's prediction $Y$, the classifier converges by identifying partitions that hold similar semantics in $U$ and grouping them to the same cluster in $Y$. The marginal entropy in Equation \ref{eq:mi v2} regularizes the classifier to ensure that each cluster is approximately assigned $n/k$ elements. On the other hand, the classifier increases its confidence on its predictions by minimizing the conditional entropy. Despite its simplicity, the experimental results presented in Section \ref{sec:results} indicate that the proposed clustering objective can effectively leverage the derived distribution $U$ to achieve high performance. \par

\subsection{Multi-Crop Strategy} 
In this work, we adopt the same multi-crop views training strategy as introduced in \cite{caron2020unsupervised} for a full exploitation of the proposed framework. Each training instance is transformed into a set of augmented views. The first two main transformed views are cropped in a negligible lower resolution of the original image, we later called this transformation as high resolution views. The additional mini-cropped views cover only a small part of the image with their resolution to be approximately the half of the original image size to reduce the convolutional operations and thus the time complexity. We named this transformation as low resolution views. During the training, $Q$ target distributions are computed only for the high resolution views, and used across all augmented views. We extend this process also to the proposed clustering objective (Eq. \ref{eq:mi v2}). Here, we compute pairwise the MI quantity based on all views; hence, if a mini-batch is transformed into two high and two low resolution views, so $V=4$, resulting in 16 combinations in total. It is found that the multi-crop strategy, as demonstrated in our empirical studies, effectively enhances the model's prediction capability.

\section{Experimental Studies}
\label{sec:results}

We evaluate the proposed clustering method, IMC-SwAV, on  four challenging colour image datasets: 1) CIFAR-10, which contains 10 equally populated classes; 2) CIFAR-100 containing 100 classes with all elements being uniform distributed. These 100 classes are also grouped into 20 super-classes with each super class consisting of five classes. For convenience, we refer to the 20 super-classes as CIFAR-100/20; 3) STL10, which contains labels only for the 13000 images and the remaining unlabeled images are from various classes. The encoder is trained across the labeled and unlabeled instances, where the classifier is trained and evaluated only for the labeled instances; and 4) Tiny-ImageNet/200, which is a subset of ImageNet containing 200 classes downsampled to a lower resolution.  Table \ref{Dataset description} presents the details of each set, including the number of training and validation elements, the number of clusters, the image resolution, and the multi-crop ranges during the training. By ``2x28+4x18", we mean two high resolution views of crop size $28$, and four low resolution views of crop size $18$. 

The proposed algorithm is compared with a range of state-of-the-art visual clustering approaches in terms of unsupervised learning metrics. We further demonstrate the effectiveness of the proposed single-phase method on datasets with up to 200 classes. Ablation studies are also performed to evaluate the effects of different hyper-parameters to examine the role of the most important components of the proposed algorithm and the sensitivity to the key parameters.

\subsection{Metrics for Unsupervised Learning}

Three quantitative  metrics for supervised learning are adopted to measure the performance of IMC-SwAV: 1) Accuracy (ACC); 2) Normalized mutual information (NMI) \cite{vinh2009information}; and 3) Adjusted rand index (ARI). The ACC reports the accuracy by finding the best mapping between the predicted labels and the ground truth labels. The mapping matrix is found using the Hungarian algorithm \cite{doi:10.1002/nav.3800020109}. NMI measures the mutual information between the two distributions (the model's prediction and the ground truth), which is scaled between zero and one. ARI measures the similarity between the two particular distributions by comparing all possible pairs and measuring those assigned in the same cluster and those in different ones.

\begin{table}[!]
\caption{Descriptions of the Datasets}
\resizebox{\columnwidth}{!}{
\begin{tabular}{lcccccccccccc}
\hline
\textbf{Name} & \begin{tabular}{@{}c@{}}\textbf{Train.} \\ \textbf{No.}\end{tabular} & \begin{tabular}{@{}c@{}}\textbf{Val.} \\ \textbf{No.}\end{tabular} & \begin{tabular}{@{}c@{}}\textbf{Classes} \\ \textbf{($\textit{k}$)}\end{tabular}& \textbf{Res.} & \begin{tabular}{@{}c@{}}\textbf{Multi-}\\ \textbf{\textbf{crops}}\end{tabular} \\ \hline
\textbf{CIFAR-10} & 50000 & 10000 & 10 & 32x32& 
\begin{tabular}{@{}c@{}} 2x28+\\4x18 \end{tabular} \\
\begin{tabular}{@{}l@{}}\textbf{CIFAR-100}\\ \textbf{\textbf{CIFAR-100/20}}\end{tabular}& 50000& 10000 & \begin{tabular}{@{}c@{}} 100\\20\end{tabular} & 32x32 & \begin{tabular}{@{}c@{}} 2x28+\\4x18 \end{tabular} \\
\textbf{STL10} & 105000 & 8000 & 10 & 
96x96 & \begin{tabular}{@{}c@{}} 2x76+\\4x52 \end{tabular} \\ 
\textbf{Tiny-ImageNet} & 100000 & 10000 & 200 & 
64x64 & \begin{tabular}{@{}c@{}} 2x56+\\4x36 \end{tabular} \\\hline
\end{tabular}
}
\label{Dataset description}
\end{table}

\begin{table*}[t]
\caption{Comparative results on the three benchmarks. The top three methods in terms of the best results highlighted. Note that our method is evaluated for $15$ independent runs across all datasets, and the average and best results are reported.}
\footnotesize
\begin{center}
\setlength\tabcolsep{1.5pt}
%\resizebox{\columnwidth}{!}{%
\begin{tabularx}{\textwidth}{l| Y Y Y|Y Y Y|Y Y Y|Y Y Y}
\hline

\multirow{2}{*}{\textbf{Method/Dataset}}& \multicolumn{3}{c}{\textbf{CIFAR-10}}  & \multicolumn{3}{c}{\textbf{CIFAR-100/20}}  & \multicolumn{3}{c}{\textbf{STL10}} & \multicolumn{3}{c}{\textbf{Tiny-ImageNet/200}}\\

\cline{2-13}
& ACC & NMI & ARI &  ACC & NMI & ARI &  ACC & NMI & ARI &  ACC & NMI & ARI \\ \hline

K-means & 22.9 & 8.7 & 4.9 & 13.0 & 8.4 & 2.8 & 19.2 & 12.5 & 6.1 & 2.5 & 6.5 & 0.5 \\

AE  \cite{NIPS2006_3048} & 31.4 & 23.9 & 16.9 & 16.5  &10.0 & 4.8 & 30.3&25.0 &16.1 & 4.1 & 13.1 & 0.7 \\

VAE \cite{Kingma2014} & 29.1 & 24.5 & 16.7 & 15.2 & 10.8 & 4.0 & 28.2 & 20.0 & 14.6 & 3.6 & 11.3 & 0.6\\

DCGAN \cite{DBLP:journals/corr/RadfordMC15} & 31.5 & 26.5 & 17.6 & 15.1& 12.0&4.5 &29.8 &21.0 &13.9 & 4.1 & 13.5 & 0.7\\

DEC \cite{pmlr-v48-xieb16} & 30.1 & 25.7 & 16.1 &  18.5&  13.6 & 5.0 & 35.9 & 27.6 & 18.6 & - & - & -\\

JULE \cite{Yang2016JointUL} & 27.2 & 19.2 & 13.8 & 13.7 & 10.3 & 3.3 & 27.7 &  18.2 & 16.4 & 3.3 & 10.2 & 0.6\\

ADC \cite{DBLP:conf/dagm/HausserPGAC18} & 32.5 & - & - & 16.0 & - & - & 53.0 & - & - & - & - & -\\

DAC \cite{DBLP:conf/iccv/ChangWMXP17} & 52.2 & 39.6 & 30.6 & 23.8 & 18.5 & 8.8 & 47.0 & 36.6 & 25.7 & 6.6 & 19.0 & 1.7 \\

IMSAT-DCGAN \cite{ntelemis2020image} & 70.0 & - & - & 32.4 & - & - & 58.7 & - & - & - & - & -\\

DDC \cite{2019arXiv190501681C} & 52.4 & 42.4 & 32.9 & - & - & - & 48.9 & 37.1 & 26.7 & - & - & -\\

DCCM \cite{DCCM} & 62.3 & 49.6 & 40.8 & 32.7 & 28.5 & 17.3 & 48.2 & 37.6 & 26.2 & \textbf{10.8} & 22.4 & 3.8\\

IIC \cite{Ji_2019_ICCV} & 61.7 & 51.1 & 41.1 & 25.7 & 22.5 & 11.7 & 59.6 & 49.6 & 39.7 & - & - & -\\

DCCS \cite{zhao2020deep} & 65.6 & 56.9 & 46.9 & - & - & - & 53.6 & 49.0 & 36.2 & - & - & -\\

PICA \cite{huang2020pica} & 69.6 & 59.1 & 51.2 & 33.7 & 31.0 & 17.1 & 71.3 & 61.1 & 53.1 & 9.8 & \textbf{27.7} & \textbf{4.0} \\

DRC \cite{zhong2020deep}  & 72.7 & 62.1 & 54.7 & 36.7 & 35.6 & 20.8 & 74.7 & 64.4 & 56.9 & - & - & -\\

EmbedUL \cite{DBLP:conf/eccv/HanPPKC20} & \textbf{81.0} & - & - & 35.3 & - & - & 66.5 & - & - & - & - & -\\

CC \cite{li2020contrastive} & 79.0 & \textbf{70.5} & \textbf{63.7} & \textbf{42.9} & \textbf{43.1} & \textbf{26.6} & \textbf{85.0} & \textbf{76.4} & \textbf{72.6} & \textbf{14.0} & \textbf{34.0} & \textbf{7.1}\\

SCAN \cite{vangansbeke2020scan} (Best) & \textbf{88.3} & \textbf{79.7} & \textbf{77.2} & \textbf{50.7 }& \textbf{48.6} & \textbf{33.3} & \textbf{80.9} & \textbf{69.8}  & \textbf{64.6} & - & - & -\\

\hline 

Supervised & 92.8 & - & - & 76.1  & - & - & 89.2 & - & - & 45.4 & - & -\\
SwAV\cite{caron2020unsupervised} + K-means & 78.4 & 67.5 & 61.3 & 40.1 & 47.0 & 10.6 & 74.9 & 70.5  & 54.0 & 18.2 & 46.8 & 5.2 \\

IMC-SwAV  (Avg$\pm$) & \textbf{89.1}$\pm$0.5 & \textbf{81.1}$\pm$0.7 & \textbf{79.0}$\pm$1.0 & \textbf{49.0}$\pm$1.8 & \textbf{50.3}$\pm$1.2 & \textbf{33.7}$\pm$1.3 & \textbf{83.1}$\pm$1.0 & \textbf{72.9}$\pm$0.9 & \textbf{68.5}$\pm$1.4 & \textbf{27.9}$\pm$0.3 & \textbf{48.5}$\pm$2.0 & \textbf{14.3}$\pm$2.1 \\

IMC-SwAV  (Best) & \textbf{89.7} & \textbf{81.8} & \textbf{80.0} & \textbf{51.9} & \textbf{52.7} & \textbf{36.1} & \textbf{85.3} & \textbf{74.7}  & \textbf{71.6} & \textbf{28.2} & \textbf{52.6} & \textbf{14.6} \\

\hline
\end{tabularx}
%}
\label{tab:Unsupervised Clustering}
\end{center}
\end{table*}

\subsection{Experimental Settings}

\subsubsection{Architecture Implementation} 
For fair comparisons, we adopt the same settings across all datasets as those given in \cite{caron2020unsupervised,pmlr-v119-chen20j}. Specifically, the proposed framework implements an encoder based on ResNet18 \cite{he2015deep} architecture. The classifier employs a multilayer perceptron net containing two hidden layers on top of the embedding layer. Similar to  \cite{caron2020unsupervised,pmlr-v119-chen20j}, we use a projection head to reduce the dimension of the embedding layer prior to the comparison with the prototypes. For the overall framework, the Adam optimizer \cite{DBLP:journals/corr/KingmaB14} is adopted for training. The learning rate is set to $5 \times 10^{-4}$ with a warmup schedule in the first $500$ training iterations. We use a decay learning rate of $0.4$ in epochs $[150, 300, 400]$ and a total of $500$ epochs is run. The mini-batch size is set to $256$ across all datasets except Tiny-ImageNet/200, for which the batch size was set $512$ due to the large number of classes in this dataset. A $\ell_2$ weight decay regularizer is used with a rate of $1\times10^{-5}$. We apply an additional two-sided regularizer, in a similar form to that reported in \cite{NEURIPS2019_ddf35421,NEURIPS2020_f3ada80d}, to the logit outputs of the classifier's representation. Specifically, we penalize any absolute value higher than five prior to the Softmax activation function, hence preventing the classifier from making predictions with a high confidence level in the early stage of the training phase as:
\begin{equation}
L^{\textit{cluster}}_\omega (U,Y) + \alpha \cdot \frac{1}{m} \sum_i^m \sum_j^k \text{max}(|x_{i,j}|-5,0) 
\end{equation}
where $x$ is the logit outputs of the classifier,  $\alpha$ is a weight parameter set to $1\times10^{-2}$ across all experiments, $m$ is the batch size, and $k$ is the number of clusters.

\subsubsection{Hyper-Parameters Selection} 
The settings below are applied across all experiments. The temperature scalar $\tau$ in Softmax smoothness is set to $0.1$ as recommended in \cite{caron2020unsupervised}. Similarly, to prevent degeneracy, we set the hyperparameter $\epsilon$ of the weighted entropy of Sinkhorn-Knopp to $0.05$ and optimize it for three iterations. The implemented prototype vectors are set to $k'= 1000$ (except in the ablation studies). In the clustering objective, we set the weight factor of the marginal entropy $\beta=4$ for all experiments.

\subsubsection{Transformation Functions} 
In this work, we follow the same transformation/augmentation scheme as presented in \cite{pmlr-v119-chen20j} across all experiments. % as in our previous studies. 
Each instance is horizontally flipped at a probability of $0.5$. All images are modified with color jittering at a rate: brightness $0.4$, contrast $0.4$, saturation $0.4$, and hue $0.2$. Color jitter is applied at a probability of $0.8$. A probability of $0.25$ is also used for a gray-scale instance transformation. The re-sizing rates are set to the range between $0.2$ and $1.0$, between $0.08$ and $0.4$, respectively, for the two main views (of high resolution) and four smaller views (of low resolution). The aspect ratio in both cases is set to $(3/4, 4/3)$. The corresponding crop sizes are given in Table \ref{Dataset description}.

\subsection{Comparative Results}

Table \ref{tab:Unsupervised Clustering} presents the results over $15$ independent runs. To demonstrate the stability of the proposed IMC-SwAV, in each run the model's parameters are randomly initialized. The same process is followed across all datasets. We present the mean, standard deviation (STD), and the best result. In contrast to the previous clustering studies in which all instances are used for training and validation, we use a similar validation process as that in \cite{10.1007/978-3-030-58607-2_16}, where the validation results are derived from the validation data that has not been seen by the model. During the training, all components of the framework are jointly trained with training instances only. We list the results obtained by a broad range of  state-of-the-art unsupervised learning algorithms for comparison. It should be stressed that the ground truth is used only for computing the relevant metrics. To further demonstrate the effectiveness of our approach, we freeze the encoder parameters and train a single layer net on top of the embedding layer in a supervised manner. We present these supervised learning result to further demonstrate the capability of the proposed unsupervised clustering approach. Finally, we compare the clustering results of K-means by using the same features extracted by the SwAV method. We also report the best results of "SwAV + K-means'' over 20 initializations of K-means' centroids. \par

As presented in Table \ref{tab:Unsupervised Clustering}, the average performance of IMC-SwAV outperforms all state-of-the-art methods on CIFAR-10, CIFAR-100/20, and Tiny-ImageNet and produces competitive results on STL10, while the best results of our method are the best among all algorithms under comparison. Note that only the best results of all algorithms under comparison are reported in Table \ref{tab:Unsupervised Clustering}. IMC-SwAV performs clustering based on the online mode in the \textit{swap} training strategy without the requirement of a pre-trained model or the implementation of multi-phase strategies, thus providing an additional advantage in terms of simplicity and training time. We observe a large margin between the clustering performance of IMC-SwAV and SwAV + K-means. Since both methods are trained on the same features, this performance improvement can be attributed to the defined clustering objective and the jointly training proposed in this work. Furthermore, the increased number of parameters of the classifier has a low computational impact during the training, with the training time on a Nvidia Quadro RTX6000 GPU being increased by $1.08\text{x}$ only. 

\begin{figure}[t]
    \centering
    \includegraphics[width=0.4 \textwidth]{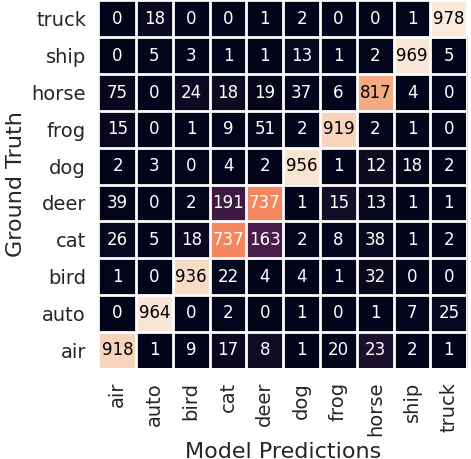}
    \caption{The above confusion matrix showing the predictions and ground truth made by the proposed model on CIFAR-10 validation set. 
    }
   \label{fig:confusion matrix}
\end{figure}

In the following, we further demonstrate our model performance by evaluating the individual classes and instances.

\subsubsection{Individual class performance} 
To examine the performance of individual class accuracies on CIFAR-10, a confusion matrix with the predictions made by IMC-SwAV on the unseen validation set is presented in Fig. \ref{fig:confusion matrix}. Note that in contrast to the previous methods, none of the implemented components are trained on the validation set. Each class consists of $1000$ instances. In the figure,  $x$-axis and  $y$-axis indicates the model's predictions and the ground truth, respectively. From these results, it can be seen that IMC-SwAV shows a high accuracy on the majority of the classes. Mismatch inaccurate predictions are mainly observed between classes `Cat' and `Deer'.

\begin{figure}[t]
    \centering
    \includegraphics[width=0.45 \textwidth]{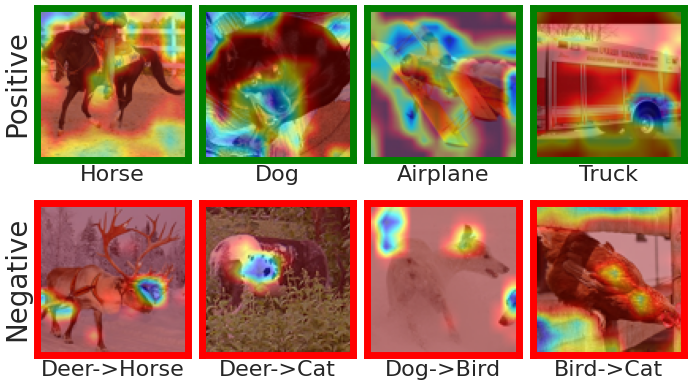}
    \caption{This illustration is an interpretation of visual activation heatmap of accurate (positive - green frame) and inaccurate (negative - red frame) prediction made by our model on STL10.
    }
   \label{fig:gradcam}
\end{figure}

\subsubsection{Visual interpretation} 
We visualize the predictions made by IMC-SwAV on STL10 dataset in Fig. \ref{fig:gradcam}. The predictions are visualized via Grad-Cam \cite{8237336}. IMC-SwAV is able to find specific visual elements to achieve the successful predictions such as airplane wings or horse's body shape. On the other hand, from the visual heatmap layers of the negative predictions, we see that the model fails to recognize specific object elements. 

\begin{figure}[t]
    \centering
    \includegraphics[width=0.45 \textwidth]{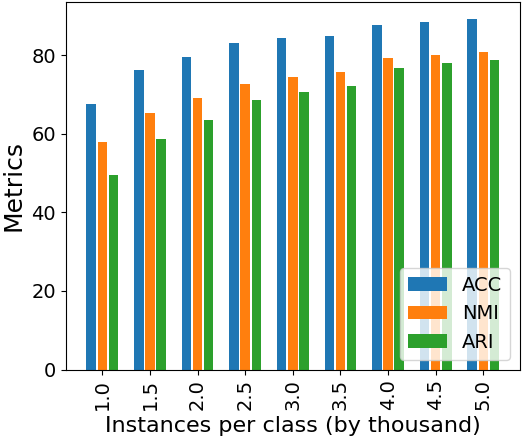}
    \caption{Performance of IMC-SwAV made on CIFAR-10 for a limited number of training elements. x-axis presents the number of training samples (by thousand) per class.
    }
   \label{fig:sample size}
\end{figure}

\begin{table}[t]
%\captionsetup{belowskip=-1pt,aboveskip=2pt}
\caption{CIFAR100 - Evaluation on 100 Classes}
\resizebox{\columnwidth}{!}{
\begin{tabular}{lcccc}
\hline
\textbf{} & \textbf{Top-1 ACC} & \textbf{Top-5 ACC}  & \textbf{NMI} & \textbf{ARI} \\ \hline
IMC-SwAV & 45.1 & 67.5 & 60.8 & 30.7 \\
SwAV + K-means & 30.2& 48.6 & 56.25 & 12.3 \\
\hline
\end{tabular}
}
%\vspace{-4mm}
\label{tab: cifar100}
\end{table}

\subsubsection{Influence of sample size}  
%To observe the robustness of our model against the changes in the number of training instances, 
We vary the number of the training instances per class by an increasing interval of $500$ per experiment within the range of $[1000,5000]$, where 5000 is the maximum number of samples in each CIFAR-10 class (recall that CIFAR-10's classes are equally spread  with each class containing $5000$ element). The model is trained on the selected subset only, and the performance of validation set is illustrated in Fig. \ref{fig:sample size}. 
%Figure \ref{fig:complex} (c) presents the results in terms of the three %performance
%metrics. 
As expected, the model's precision is gradually decreasing when fewer training samples are considered. Notably, IMC-SwAV's performance remains stable when the sample size is larger than $2500$.

\subsubsection{Effectiveness to a large number of clusters} 
%CIFAR100-20 contains $20$ super-classes, with each consisting of five sub-classes. Hence, an additional annotation of a total $100$ classes is provided. 
To further validate our method on a dataset with a large number of clusters, we conduct an experiment on the 100 sub-class of CIFAR-100 to  evaluate the clustering performance of IMC-SwAV. The features extracted by IMC-SwAV are directly compared with those obtained by the K-means algorithm. The metrics of Top-1 and Top-5 accuracy, as well as the rest unsupervised metrics are listed in Table \ref{tab: cifar100}. 
These results demonstrate that IMC-SwAV performs well on this highly demanding clustering task. The proposed algorithm achieves a large margin of $15\%$ on Top-1 ACC and $19\%$ on Top-5 ACC. %This might be attributed to the implemented classifier and to the proposed training objective.

\subsection{Ablation Studies}

In this subsection, we experiment with a variety of hyper-parameters to evaluate the effectiveness of the proposed IMC-SwAV. All parameters remain unchanged, as per the main settings, except for those to be studied. \par

\begin{figure}[t]
    \centering
    \includegraphics[width=0.45 \textwidth]{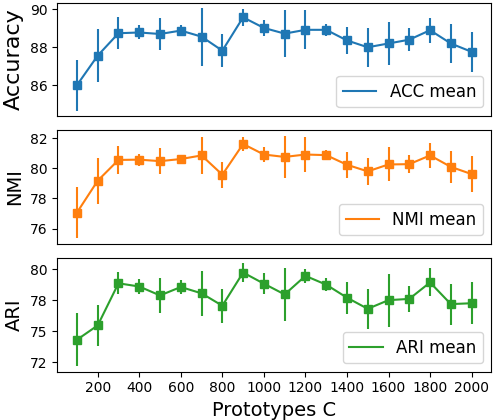}
    \caption{These diagrams plot the performance metrics on CIFAR-10 over different numbers of prototypes. From top to low: ACC, NMI and ARI. All figures report the number of prototypes in x-axis.
    }
   \label{fig:prototypes illustration}
\end{figure} 

\begin{table}[t]
\caption{IMC-SwAV with different ResNet architectures indicated by the method suffix}
\setlength\tabcolsep{3.5pt}
\begin{tabularx}{250.pt}{l|YYY|YYY|YYY}
\hline

\multirow{2}{*}{\begin{tabular}{@{}c@{}}\end{tabular}}& \multicolumn{3}{c}{\textbf{CIFAR-10}}  & \multicolumn{3}{c}{\textbf{CIFAR-100/20}}  & \multicolumn{3}{c}{\textbf{STL10}} \\

\cline{2-10}
& ACC & NMI & ARI &  ACC & NMI & ARI &  ACC & NMI & ARI \\ \hline

IMC-SwAV-18 (Avg) & 89.1  & 81.1  & 79.0 & 49.0 & 50.3 & 33.7 & 83.1  & 72.9  & 68.5 \\
IMC-SwAV-18 (Best) & 89.7 & 81.8 & 80.0 & 51.9 & 52.7 & 36.1 & 85.3 & 74.7 & 71.6 \\

IMC-SwAV-34 (Avg) & 89.5 & 81.7 & 79.9 & 50.2 & 51.2 & 34.6  & 84.5 & 74.8  & 70.9 \\
IMC-SwAV-34 (Best)  & 90.2 & 82.4 & 80.9 & 52.1 & 53.2 & 36.3 & 86.0 & 76.7  & 73.3 \\
IMC-SwAV-50 (Avg) & 91.0 & 83.8 & 82.6  & 51.2  & 52.6 & 35.6 &  86.3 & 77.3  & 73.8 \\
IMC-SwAV-50 (Best) & 91.4 & 84.1 & 82.9 & 52.7 & 54.0 & 37.0 & 87.1 & 77.9  & 75.0 \\
\hline
\end{tabularx}
\label{tab: effects resnet}
\end{table}

\subsubsection{Prototypes} 
We examine the performance of IMC-SwAV by varying the number of prototypes, $k'$, in the range of $[100,2000]$ while keeping all other parameters unchanged. Figure \ref{fig:prototypes illustration} shows the results from five independent runs of the unsupervised learning task. From these results, we conclude that IMC-SwAV shows robustness over different numbers of prototypes with the lowest mean value above $86.5\%$. We also note that for $k'$ larger than $300$, all three performance metrics become less sensitive to $k'$. Additionally, we note that our choice to set $k'$ to $1000$ prototypes in the main settings is not the optimum in terms of all metrics on CIFAR-10. However, in unsupervised learning, the ground truth is unknown and hence it is more realistic to compare the performance without using the optimal parameter setting.

\subsubsection{Experiment with different encoder architectures}  
In this subsection, we examine the performance of the proposed IMC-SwAV by varying the encoder's architecture by using different types of the ResNet \cite{he2015deep} models. Specifically, we compare the performance of the proposed method when ResNet34 (21.3M parameters) and ResNet50 (23.5M parameters), respectively, is adopted to replace  ResNet18 (11.1M parameters) used in the main experiments. All models are evaluated over five independent runs (the results of ResNet18 are averaged over 15 runs as in the main experiment) and the average and best recorded performances are reported for each model. The reported metrics are on the unseen test subset, and all frameworks are optimised with training subset only. The results are listed in Table \ref{tab: effects resnet}. As expected, larger models achieve slightly better performance in terms of all metrics. Note also that the discrepancies between different metrics of each model are minor, with the highest differences being observed difference on CIFAR-10. %This demonstrates the performance level and robustness is due to the IMC-SwAV method introduced here

\subsubsection{Multiple Crops} 
We demonstrate the impact of the multiple crops strategy on the clustering performance on CIFAR-10. We change the number of low resolution cropped views with all other settings unchanged. First, a basic setting with two main views of the original image dimensions is evaluated. Then, we add two, four, and six additional low resolution views, respectively. The size of the two main views is decreased to $28\text{x}28$ to reduce the computation time. The resizing ratio remains unchanged: $[ 0.2,1.0]$ for the two main views, and $[0.08,0.4]$ for the low resolution views. We report the mean and STD of the metrics over five independent experiments. Additionally, the accuracy of a single layer net, which is trained independently on generated features through a supervised mode, is also included. \par

\begin{table}[t]
%\captionsetup{belowskip=-5pt,aboveskip=0pt}
\caption{Effectiveness of Multiple Crops Strategies}
\resizebox{\columnwidth}{!}{
\begin{tabular}{l|ccccc}
\hline
\textbf{Crops} & \textbf{ACC} & \textbf{NMI} & \textbf{ARI} & \textbf{Sup.}& \begin{tabular}{@{}c@{}}\textbf{Train} \\ \textbf{Time}\end{tabular} \\ \hline
2x32 & 78.7$\pm$1.0 & 66.2$\pm$1.1 & 61.2$\pm$1.3 & 86.1 & $1\text{x}$  \\

2x28+2x18 & 87.2$\pm$1.1  & 78.6$\pm$1.4 & 75.8$\pm$2.0 & 92.1 & $1.1\text{x}$ \\

2x28+4x18 & 89.1$\pm$0.5  & 81.1$\pm$0.7 & 79.0$\pm$1.0 & 92.8 & $1.7\text{x}$ \\

2x28+6x18 & 89.4$\pm$0.6 & 81.2$\pm$0.9  & 79.0$\pm$1.8 & 93.2 & $2.4\text{x}$  \\
\hline
\end{tabular}
}
\label{tab: effects multiple}
\end{table}

\begin{figure}[t]
    \centering
    \includegraphics[width=0.45 \textwidth]{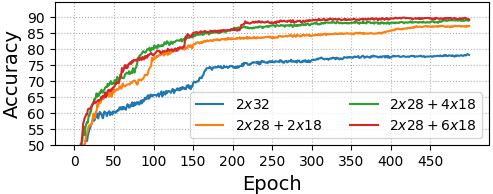}
    \caption{Illustration presents the accuracy per epoch of each multi-crop implementation of Table \ref{tab: effects multiple}.
    }
   \label{fig:training epochs}
\end{figure}

\begin{figure}[t]
    \centering
    \includegraphics[width=0.45 \textwidth]{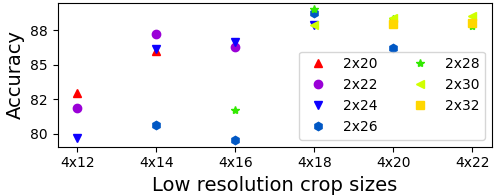}
    \caption{Presentation with various combinations of high and low crop sizes. $x\text{-axis}$ represent different low resolution crop sizes and high resolution are associated with different colours and symbols. 
    }
   \label{fig:views resolution}
\end{figure}

Table \ref{tab: effects multiple} presents the results of the multiple crop experiments. From results, we notice that the additional low resolution views increase the performance in terms of all metrics, for both the unsupervised and supervised models. The last column presents the ratio of training time based on a basic training strategy of $2\text{x}32$. The training accuracy over the iterations are presented in Fig. \ref{fig:training epochs} for each experiment. We note that the mini-crop strategy of ``$2\text{x}28\text{+}4\text{x}18$" already reaches satisfactory results ($87.7$ ACC) within $250$ iterations, in comparison to the full experiment of a simple setup. Recall, the two-crop strategy holds an advantage by using the original resolution of $2\text{x}32$ instead of $2\text{x}28$.\par

\subsubsection{Combinations of crop sizes} 
To further examine the impact of multi-crop's strategy, we vary the resolution of crop sizes of main views (high) and smaller views (low) within the intervals of $[20,32]$ and $[12,22]$, respectively. All experiments are performed on CIFAR-10 based on the main setting. We also preserve the same crop ratios and the number of views (two main views and four smaller views). Since, the resizing ratio of the smaller view is set to $0.4$, the difference between high and low resolutions does not exceed the absolute difference of the maximum $14$ pixels and minimum four pixels, to keep proportional to the crop size of main view. Figure \ref{fig:views resolution} presents our findings in terms of ACC. The performance of IMC-SwAV slightly degrades in a very low resolution in combination with ``$2\text{x}20\text{+}4\text{x}12$", probably because the model fails to capture sufficient semantic details. On the other hand, all combinations above the crop sizes of $\text{(high view)} \geq 26$ and $\text{(low view)}\geq18$ exhibit a satisfactory accuracy above $86\%$.

\subsubsection{Other representation learning strategies}  
In this section, we aim to demonstrate that our clustering strategy is not restricted to the SwAV \cite{caron2020unsupervised} framework and is also applicable to other self-supervised learning strategies. Here, we evaluate our single-phase clustering method that adopts SeLa \cite{asano2020self}, another grouping based self-supervised learning method, to replace SwAV. Similar to IMC-SwAV, the components of the encoder and the classifier are jointly trained in a single-phase mode. We compare the performance of the variant of our clustering method using Sela (called IMC-SeLa) with the K-means algorithm, and a supervised classifier net trained on same extracted features. The parameters of the encoder model are optimized with the self-supervised loss function of the adopted strategy only without using any additional loss functions or pre-trained models. Similar to the main experiments, the framework is optimized on training subset and the performance evaluation is made on the validation subset. Each method is performed for five independent runs and the best result is reported. For a fair comparison, the architecture is similar to the main experiment with the encoder being based on a ResNet18 network. \par

The comparative results are listed in Table \ref{tab: sela}, from which a performance degradation of all methods under comparison can be observed, implying that the extracted features are less separable. This can be easily validated by examining the results of the supervised linear classifier net (Supervised in the Tables) trained on the extracted features (for SwAV \cite{caron2020unsupervised} in Table \ref{tab:Unsupervised Clustering} and for SeLa \cite{asano2020self} in Table \ref{tab: sela}). This can also be confirmed by the K-means clustering approach on the same features extracted by the two self-supervised techniques. 

These results indicate that the final performance of the proposed framework also depends on the performance of the self-supervised learning algorithm it adopts. Nevertheless, IMC-SeLa still outperforms "SeLA + K-means'' at a large margin on CIFAR-10, CIFAR-100/20 and Tiny-ImageNet/200. 

\begin{table}[t]
\caption{Clustering strategy adopted to SeLA training}
\setlength\tabcolsep{3.5pt}
\begin{tabularx}{250.pt}{l|YYY|YYY|YYY}
\hline

\multirow{2}{*}{\begin{tabular}{@{}c@{}}\textbf{Method/}\\ \textbf{Dataset.}\end{tabular}}& \multicolumn{3}{c}{\textbf{CIFAR-10}}  & \multicolumn{3}{c}{\textbf{CIFAR-100/20}}  & \multicolumn{3}{c}{\textbf{Tiny-ImgNet200}} \\

\cline{2-10}
& ACC & NMI & ARI &  ACC & NMI & ARI &  ACC & NMI & ARI \\ \hline

IMC-SeLa & 74.5 & 65.2 & 59.1 & 39.4 & 40.7 & 25.1 & 19.1 & 45.2 & 7.9 \\
SeLA \cite{asano2020self} + Kmeans  & 64.0 & 56.4 & 47.3 & 34.7 & 34.3 & 18.0 & 17.1 & 43.2  & 6.7 \\
Supervised  & 87.1 & - & - & 66.6 & - & - & 34.2 & -  & - \\
\hline
\end{tabularx}
%}
\label{tab: sela}
\end{table}

\section{Conclusion}

This study presents a single-phase framework for image clustering, called IMC-SwAV.  We introduce a modified mutual information to jointly train the framework by maximizing the dependency of a self-labelling assignment strategy and an implemented classifier. IMC-SwAV achieves highly competitive performance on challenging datasets compared to the state-of-the-art. In addition to the encouraging clustering performance, we demonstrate the robustness of the proposed method to the parameter settings and the possibility to adopt a different self-supervised learning technique. \par

Although the proposed model achieves significant capabilities in clustering, the multi-crop training strategy adopted in the framework slightly increases the time complexity. Additionally, we observe a dependency of its performance on the grouping based self-supervised learning method. Our future work will aim to remove the requirement of using the multi-crop views without deteriorating the performance of the overall framework. Moreover, we will examine the effectiveness of other grouping based self-supervised learning algorithms in the proposed single-phase training framework to make it available for a wider range of applications.

{\small
\bibliographystyle{ieee_fullname}

}

\end{document}